\documentclass{article}
\usepackage{ijcai24}

\usepackage{times}
\usepackage{soul}
\usepackage{url}
\usepackage[hidelinks]{hyperref}
\usepackage[utf8]{inputenc}
\usepackage[small]{caption}
\usepackage{graphicx}
\usepackage{amsmath}
\usepackage{amsthm}
\usepackage{booktabs}
\usepackage{makecell}
\usepackage{caption}
\usepackage{subcaption}
\usepackage{amsfonts}
\usepackage{algorithm}
\usepackage{algorithmic}
\usepackage{bbm}
\usepackage[switch]{lineno}

\title{MultifacetEval: Multifaceted Evaluation to Probe LLMs in  \\   Mastering Medical Knowledge}

\author{
Yuxuan Zhou
\and
Xien Liu\thanks{Corresponding author}\and
Chen Ning\And
Ji Wu
\affiliations
Department of Electronic Engineering, Tsinghua University, Beijing, 100084, China
\emails
\{zhou-yx21, nc22\}@mails.tsinghua.edu.cn,
\{xeliu, wuji\_ee\}@mail.tsinghua.edu.cn
}

\begin{document}

\maketitle

\begin{abstract}

Large language models (LLMs) have excelled across domains, also delivering notable performance on the medical evaluation benchmarks, such as MedQA. 
However, there still exists a significant gap between the reported performance and the practical effectiveness in real-world medical scenarios. 
In this paper, we aim to explore the causes of this gap by employing a multifaceted examination schema to systematically probe the actual mastery of medical knowledge by current LLMs. Specifically, we develop a novel evaluation framework \textbf{MultifacetEval} to examine the degree and coverage of LLMs in encoding and mastering medical knowledge at multiple facets (comparison, rectification, discrimination, and verification) concurrently.
Based on the MultifacetEval framework, we construct two multifaceted evaluation datasets: MultiDiseK (by producing questions from a clinical disease knowledge base)  and  MultiMedQA (by rephrasing each question from a medical benchmark MedQA into multifaceted questions).
The experimental results on these multifaceted datasets demonstrate that the extent of current LLMs in mastering medical knowledge is far below their performance on existing medical benchmarks, suggesting that they lack depth, precision, and comprehensiveness in mastering medical knowledge. Consequently, current LLMs are not yet ready for application in real-world medical tasks.
The codes and datasets are available at \url{https://github.com/THUMLP/MultifacetEval}.

\end{abstract}

\section{Introduction}
The rapid advancement of large language model (LLM) technology has achieved great success in various domains \cite{romera2023mathematical,madani2023large,boiko2023autonomous}. Current LLMs encode extensive knowledge through pretraining on massive unlabeled data. Some are further finetuned on supervised datasets to be adapted to specific downstream tasks. Recently, famous general LLMs like GPT-4 \cite{openai2023gpt4} and Gemini-pro \cite{team2023gemini}, as well as medical-domain-specific LLMs such as Med-PaLM \cite{singhal2023large}, are reported to have encoded vast medical knowledge and achieved significant performance on several medical benchmarks, surpassing previous state-of-the-art models by a considerable margin \cite{kung2023performance,nori2023capabilities,nori2023can}. Nevertheless, despite their impressive performance on existing benchmarks, these LLMs still face challenges in addressing real-world medical problems \cite{thirunavukarasu2023large,clusmann2023future,wornow2023shaky}. This leads to a significant gap between evaluation results and practical performance in the medical domain. Therefore, in this paper, we aim to study the underlying causes of this gap by systematically investigating the \textbf{depth} of medical knowledge mastery in current LLMs.

\begin{figure}[t]
    \centering
    \includegraphics[width=0.47\textwidth]{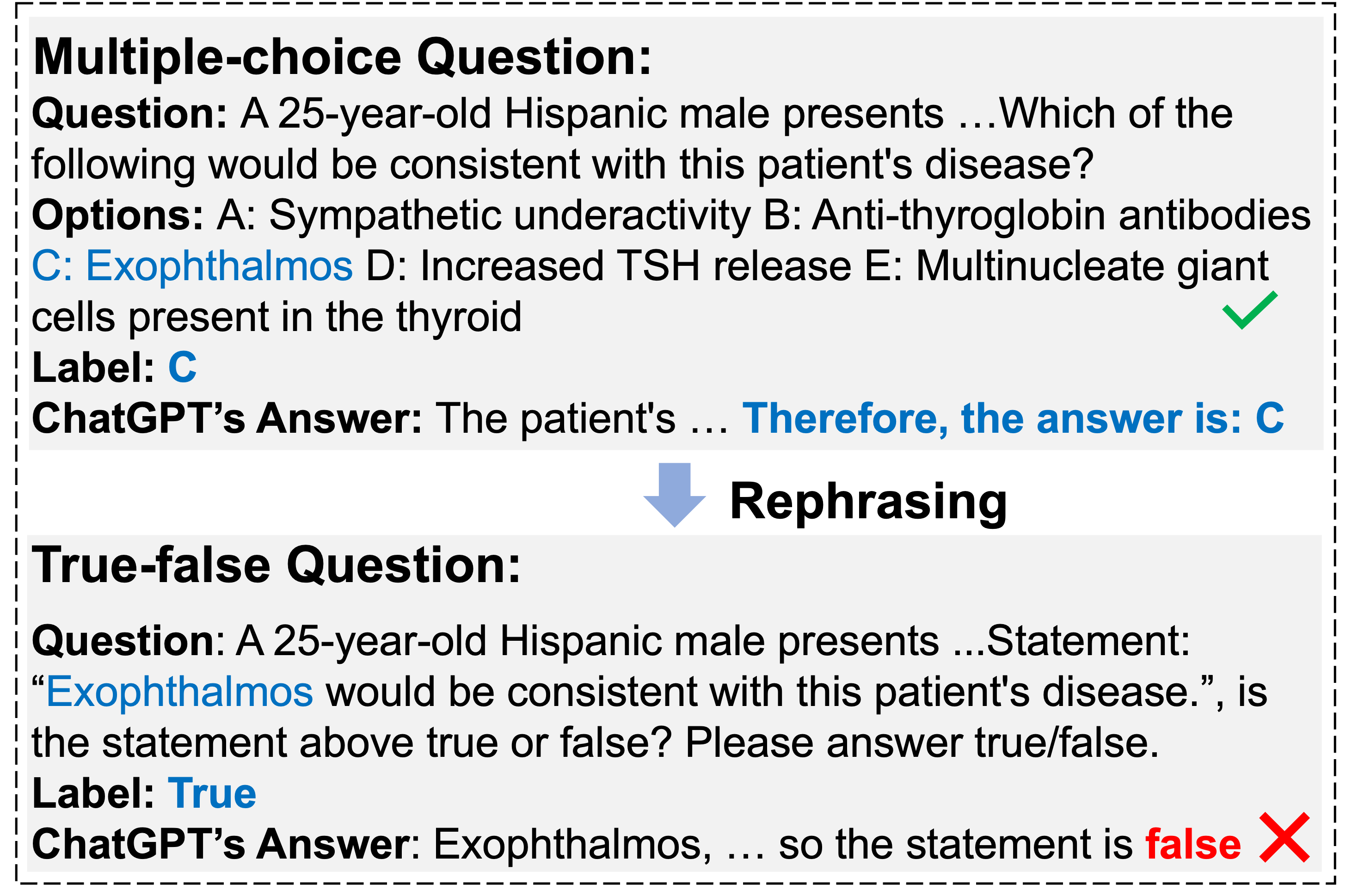}
    \caption{GPT-3.5-turbo responding to medical exam problems assessing the same knowledge point but in different formats.}
    \label{fig:example}
\end{figure}
Several medical benchmarks have been proposed to measure LLMs' capacities in the medical domain. Most current medical benchmarks assess LLMs by medical question-answering tasks \cite{jin2021disease,medmcqa,hendrycks2020measuring,jin-etal-2019-pubmedqa,LiveMedQA2017,MEDINFO19,singhal2023large}. Other benchmarks also evaluate LLMs in the forms of medical dialogue \cite{zeng2020meddialog} or other traditional NLP tasks (e.g., relation extraction, NER) based on medical corpora \cite{zhang2022cblue}. Nevertheless, most existing medical benchmarks rely on a specific question type (e.g., multiple-choice questions) to evaluate LLMs. Therefore, they may overestimate the performance of current LLMs, as certain LLMs may have been finetuned for specific question types. Consequently, their performance on specific question types would significantly surpass those on other questions. Even if some benchmarks evaluate LLMs' medical capabilities from various facets, each facet is evaluated based on distinct sets of knowledge points. Therefore, the outcomes of these benchmarks still cannot reflect LLMs' mastery on the same knowledge points across diverse facets. Meanwhile, we found it necessary to conduct \textbf{multifaceted} evaluation on the same knowledge point. Figure \ref{fig:example} illustrates GPT-3.5-turbo's response to two medical exam problems assessing the same knowledge point but in different question types. The multiple-choice question is extracted from the United States Medical Licensing Examination (USMLE). In contrast, the true-false question is adapted from the original question by substituting the phrase ``Which of the following" with the correct option, evaluating LLMs' ability to verify statements based on corresponding medical knowledge. Although GPT-3.5-turbo successfully chooses the symptom of the patient's disease, it judges the statement that is consistent with MCQ's answer as false, conflicting with its previous prediction. This highlights the importance of conducting multifaceted evaluations on identical medical knowledge points for a systematically analysis of LLMs' knowledge mastery. 

\begin{figure}[t]
    \centering
    \includegraphics[width=0.47\textwidth]{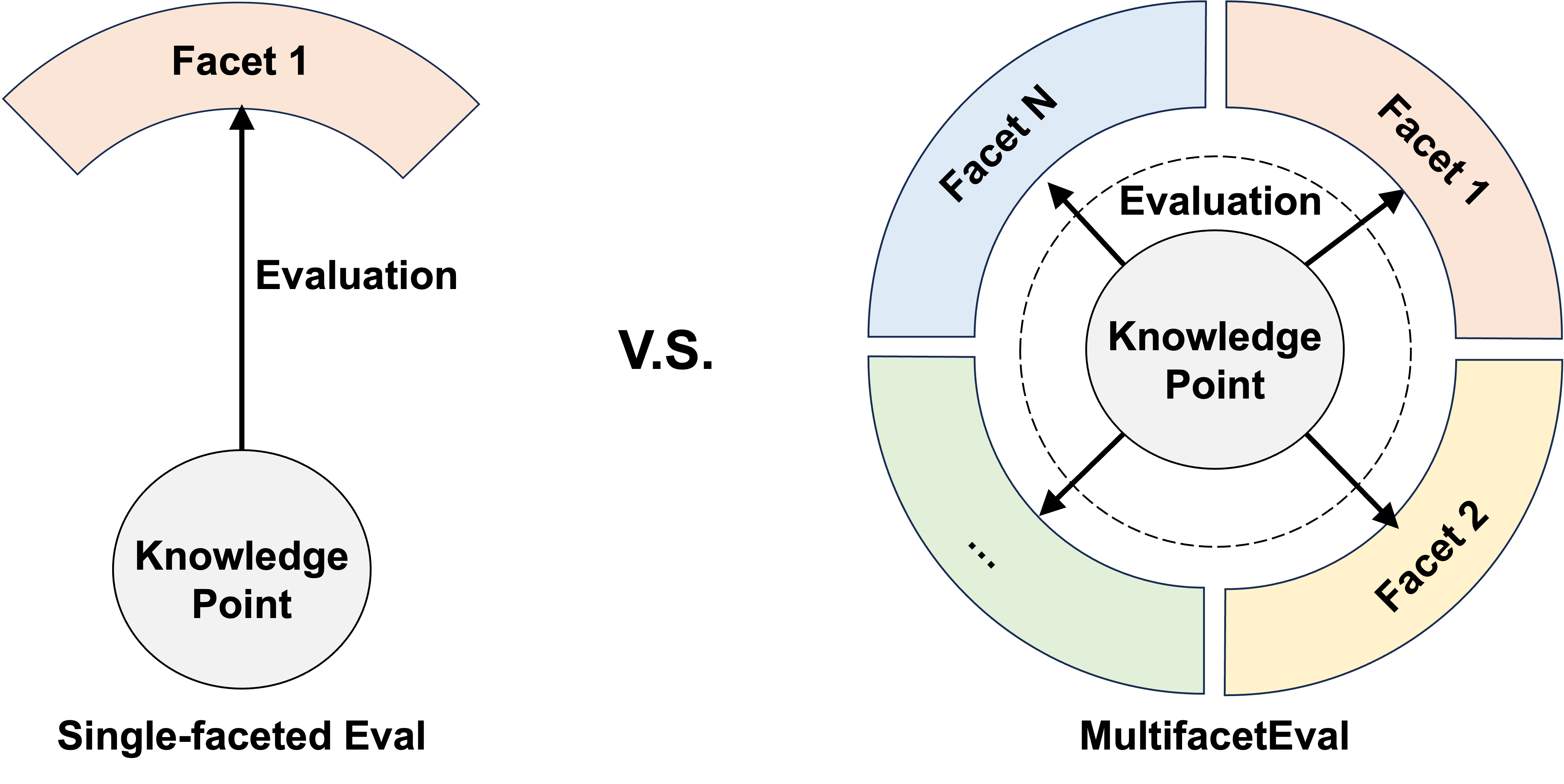}
    \caption{Principle of the proposed multifaceted evaluation.}
    \label{fig:multifaceted}
\end{figure}
In contrast to existing evaluation benchmarks, \textit{current education systems generally utilize various assessment methods, including assignments, quizzes, projects, and exams, to evaluate students' comprehensive mastery of the same knowledge point from multiple facets}. Inspired by this, we propose a novel multifaceted evaluation approach \textbf{MultifacetEval} to evaluate the actual medical knowledge mastery of current LLMs from multiple facets. Figure \ref{fig:multifaceted} illustrates the principle of this approach. Specifically, we generate a series of questions for each knowledge point of interest with various question types. These questions emphasize evaluating this knowledge point from different facets, including comparison, discrimination, verification, and rectification capabilities. Therefore, the proposed approach would provide a more comprehensive evaluation of LLMs' medical knowledge mastery compared with conventional medical benchmarks that rely on a single evaluation facet. The proposed approach also possesses strong versatility, as it can generate multifaceted questions by directly crafting them based on knowledge points in medical knowledge bases or rephrasing questions on existing medical evaluation benchmarks. 

To validate the effectiveness of the proposed multifaceted evaluation method, we apply the proposed method to construct two new evaluation datasets based on a medical knowledge database and a medical benchmark MedQA \cite{jin2021disease}, respectively. 
A total of 13 well-known general and medical LLMs are evaluated on these datasets. The experimental results indicate that current LLMs lack a comprehensive, precise, and in-depth mastery of medical knowledge despite their considerable performance on existing medical benchmarks. Moreover, the results demonstrate that current LLMs possess excellent comparison capability, while they have not well mastered other capabilities such as discrimination, verification, and rectification in the medical domain. Our contributions can be summarized as follows:
\begin{itemize}
    \item We propose a novel multifaceted evaluation schema (MultifacetEval) to evaluate LLMs' medical knowledge mastery on the same knowledge point from various facets instead of a single facet in existing benchmarks. The proposed method can more accurately evaluate the LLMs' mastery of medical knowledge.
    \item Based on the proposed method, we generate two novel multifaceted datasets, \textbf{MultiDiseK} and \textbf{MultiMedQA}, based on a medical knowledge base and a well-known medical benchmark MedQA, respectively. The performance of these two datasets can more comprehensively reflect LLMs' mastery of medical knowledge.
    \item The experimental results reveal that the genuine extent of medical knowledge mastery in current LLMs is significantly lower than that evaluated by existing medical benchmarks. Additionally, we observe substantial variations in LLMs' performance across different facets.
\end{itemize}
\begin{figure*}[t]
    \centering
    \includegraphics[width=1\textwidth]{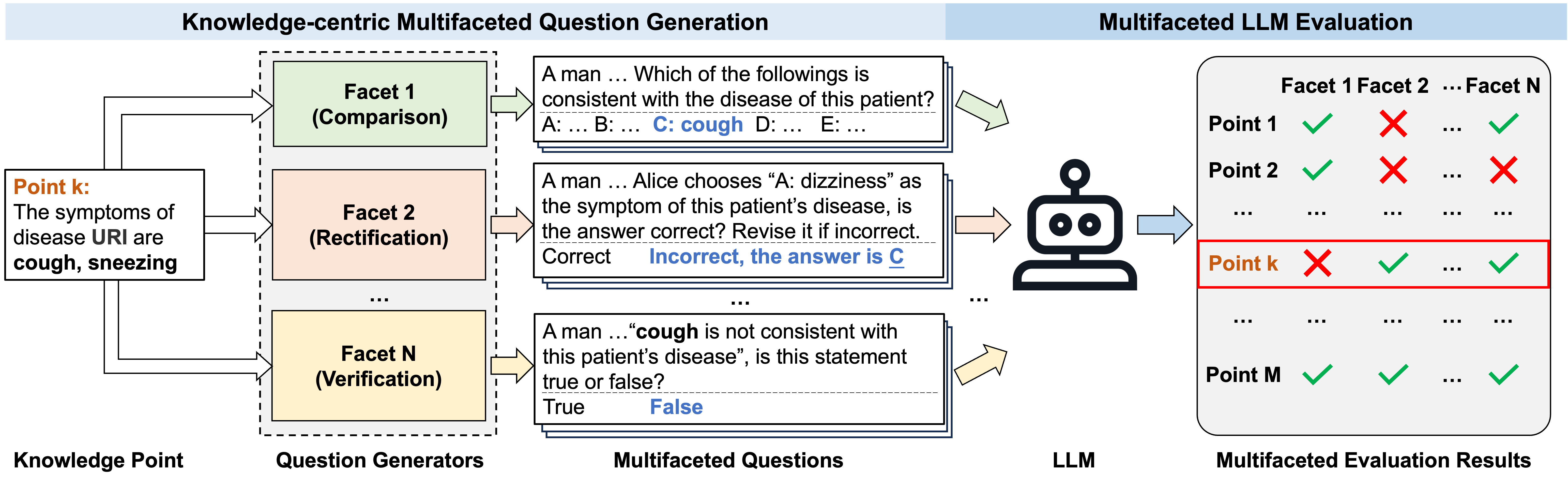}
    \caption{Framework of the proposed multifaceted evaluation approach that evaluates LLMs' medical knowledge mastery from various facets.}
    \label{fig:method}
\end{figure*}
\section{Related Work}
\paragraph{Large Language Models on Medical Tasks}
Recently, some famous LLMs are reported to encode medical knowledge and achieve considerable performance on existing medical benchmarks. General LLMs such as Flan-PaLM and GPT-4 are reported to achieve state-of-the-art performance on multiple datasets \cite{singhal2023large,nori2023can}. For example, they achieve accuracies of 67.6 and 90.2 on a medical exam benchmark MedQA \cite{jin2021disease}, largely surpassing the prior SOTA models. Several LLMs specially pretrained or finetuned on the medical corpora, such as Med-PaLM, Med-PaLM2 \cite{singhal2023towards}, ClinicalCamel \cite{toma2023clinical}, and Med42 \cite{med42}, are also proposed to address problems in the medical domain and achieve high performance on various medical benchmarks. However, these models cannot tackle problems in real medical scenarios. Our study aims to investigate the gap between the high evaluation performance and the limited practical effectiveness of existing LLMs in the medical domain.
\paragraph{Medical Evaluation Benchmarks}
Current medical evaluation benchmarks can be classified into three classes: (1) Question-answering datasets with problems collected from different sources, including medical exams \cite{jin2021disease,medmcqa,hendrycks2020measuring}, scientific literature \cite{jin-etal-2019-pubmedqa}, and consumer health questions \cite{LiveMedQA2017,MEDINFO19,singhal2023large}; (2) medical dialogue datasets \cite{zeng2020meddialog,yang2020generation}; (3) datasets \cite{peng2019transfer,zhang2022cblue} involving conventional NLP tasks (NER, relation extraction, NLI) on medical corpora. Some of these datasets assess LLMs from a single facet. Others evaluate LLMs with multiple tasks, while the tasks are constructed on different groups of knowledge points. In this paper, we design a new evaluation method to evaluate LLMs' mastery of the same knowledge point from multiple facets.
\section{Knowledge-Centric Multifaceted Evaluation}
\subsection{Multifaceted Evaluation Schema}\label{sec:schema}
The proposed multifaceted evaluation approach is motivated by the current education systems, where various assessment methods, including assignments, projects, and exams, are employed to comprehensively evaluate whether students have truly mastered a particular knowledge point. Given a knowledge point $\mathrm{k}$ and a series of $N$ evaluation facets $\mathbf{f}=[f^1,f^2,\cdots,f^N]^{\textbf{T}}$, the performance of an LLM ($\mathrm{M}$) evaluated through the proposed multifaceted evaluation is:
\begin{equation}
    \mathbf{f}_\mathrm{k}(\mathrm{M}) = [f^1_\mathrm{k}(\mathrm{M}),f^2_\mathrm{k}(\mathrm{M}),\cdots,f^N_\mathrm{k}(\mathrm{M})]^{\textbf{T}}
\end{equation}
Where $f^i_\mathrm{k}$ denotes the specific questions designed to emphasize the evaluation of the $i^{th}$ facet related to the knowledge point $\mathrm{k}$, and $f^i_\mathrm{k}(\mathrm{M})\in\{0,1\}$ is the evaluation outcome: $f^i_\mathrm{k}(\mathrm{M})=1$ if all questions in $f^i_\mathrm{k}$ are answered correctly, and 0 otherwise. Compared with single-faceted evaluation, the proposed multifaceted evaluation method conveys more comprehensive information about the mastery of a specific knowledge point. The proposed multifaceted evaluation schema demonstrates strong \textbf{transferability}: it can be applied in other domains by adjusting evaluation facets and question generation strategies.
\subsection{Facets of Medical Knowledge Mastery}\label{sec:facets}
We employ the proposed evaluation schema to systematically probe current LLMs in mastering medical knowledge. Considering the characteristics of medical scenarios, we set $N=4$ and design a total of four evaluation facets of capabilities that are essential for solving real medical problems:\\
\textbf{Comparison} ($f^1$): The ability to compare different medical entities/events and choose the most suitable one that meets some criteria. It is crucial for medical applications such as diagnosis and drug recommendation.\\
\textbf{Rectification} ($f^2$): The capability to identify errors in the medical process (treatment, diagnosis) and offer corresponding corrections. Rectification plays an important role in medical scenarios such as computer aided diagnosis.\\
\textbf{Discrimination} ($f^3$): The capacity to recognize and differentiate between medical concepts accurately. Discrimination of medical concepts is the bedrock of medical applications such as clinical decision support and personalized medicine.\\
\textbf{Verification} ($f^4$): The ability to determine the veracity of a statement based on the acquired knowledge. Such capability is highly demanded in the quality assessment of electronic health records and laboratory results.

\subsection{Multifaceted Medical Evaluation Framework}\label{sec:framework}
Built on the facets discussed above, we design a multifaceted medical evaluation framework to comprehensively evaluate mastery of medical knowledge by LLMs from these evaluation facets. Figure \ref{fig:method} illustrates an overview of the proposed multifaceted evaluation framework. Given a set of medical knowledge points $\mathcal{K}$, the framework evaluates LLMs' mastery of medical knowledge through two steps:
\paragraph{Multifaceted Question Generation} In the first step, we generate multiple questions from diverse evaluation facets for each knowledge point in the set: $\mathrm{k}\to\{f^i_{\mathrm{k}}|1\leq i\leq N\}$, $\text{where } \mathrm{k}\in\mathcal{K}$. Specifically, we design four question types, including multiple-choice questions, revision questions, multiple-answer questions, and true-false questions, to emphasize the evaluation of the comparison, rectification, discrimination, and verification facets, respectively:\\
\textbf{(1) Multiple-Choice Questions}: We maintain the multiple-choice questions (MCQ) applied in existing benchmarks to emphasize the evaluation of the comparison facet. A multiple-choice question comprises a question and multiple options (typically 4). To answer multiple-choice questions accurately, participants must compare the given options and select the most suitable choice that fits the question.\\
\begin{figure}[t]
    \centering
    \includegraphics[width=0.47\textwidth]{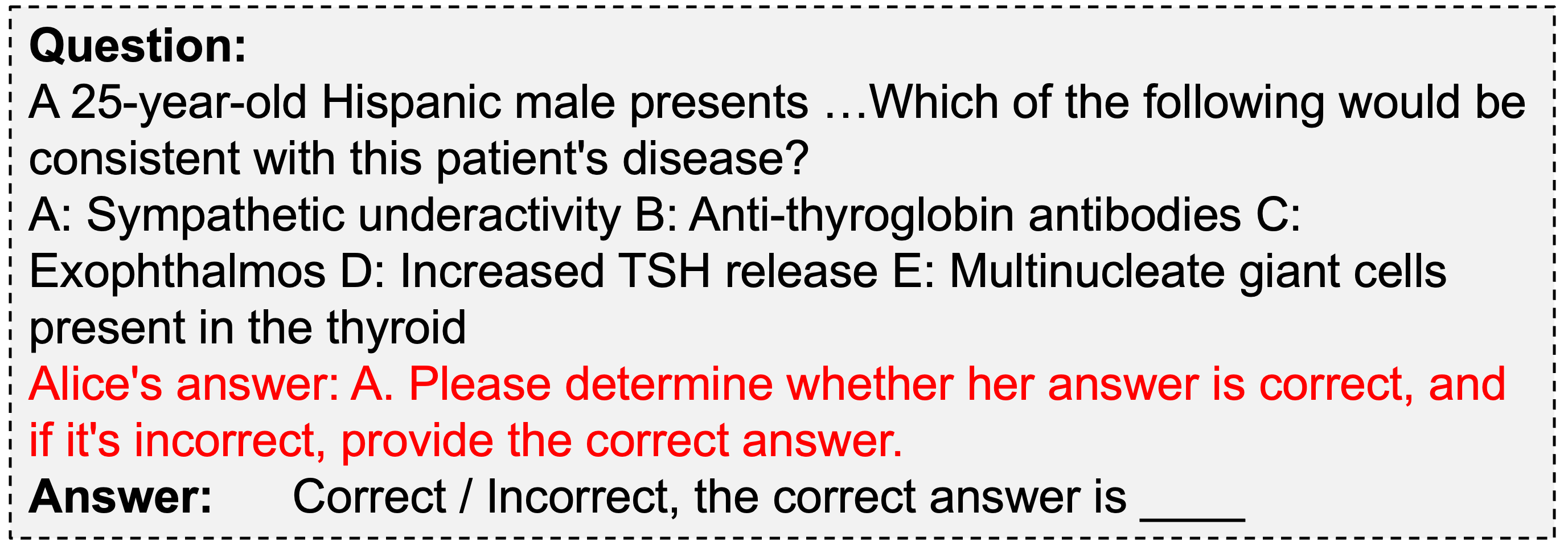}
    \caption{Example of the proposed revision question.}
    \label{fig:RQ}
\end{figure}
\textbf{(2) Revision Questions}: We design a new question type named ``revision" question (RQ) to focus on evaluating the rectification capabilities of LLMs. Figure \ref{fig:RQ} illustrates an example of this question type. A revision question comprises a multiple-choice question and a \textbf{provided option} (not necessarily correct) to this question. Participants are asked to recheck the correctness of the given option based on the question, and revise the answer appropriately if needed.\\
\textbf{(3) Multiple-Answer Questions}: 
We consider leveraging multiple-answer questions (MAQ) to highlight the evaluation of the discrimination capability. In contrast to MCQs, a multiple-answer question consists of several options, with one or more aligning with the given question. Effectively answering MAQs requires a comprehensive and precise mastery of discriminative knowledge for all options, as MAQ answers cannot be determined through option-wise comparison.\\
\textbf{(4)True-false Questions}: We utilize true-false questions (TFQ) to emphasize the assessment of the verification facet. A true-false question generally presents a statement that can be verified based on the corresponding medical knowledge and information provided in the statement. True-false questions do not include options that may provide clues, requiring ones to mastery medical knowledge accurately.
\paragraph{Multifaceted LLM Evaluation} In the next step, we evaluate the LLM $\mathrm{M}$ with the generated multifaceted questions to obtain comprehensive evaluation results on each knowledge points:
$
    f^i_\mathrm{k} \to f^i_\mathrm{k}(\mathrm{M})
$
$\text{where } k\in \mathcal{K}\text{ and }1\leq i \leq N$. Finally, the proposed evaluation framework produces comprehensive evaluation outcomes for $\mathrm{M}$ across all the knowledge points: $\{\mathbf{f}_{\mathrm{k}}(\mathrm{M})|\mathrm{k}\in\mathcal{K}\}$. To reflect the LLM's comprehensive mastery of individual knowledge points, we define a knowledge point $\mathrm{k}$ is \textbf{mastered} by $\mathrm{M}$ under facets $\{f^1, f^2...,f^N\}$, if the function
\begin{equation}
    r_\mathrm{k}(\mathrm{M})=\prod_{i=1}^{N}f^i_\mathrm{k}(\mathrm{M})
\end{equation}
equals 1. Here, $r_\mathrm{k}(\mathrm{M})=1$ only when $f^i_\mathrm{k}(\mathrm{M})=1$ holds for $1\leq i \leq N$, indicating accurate answers to all questions from these facets. The overall performance can therefore be represented as the \textit{proportion of mastered knowledge points}: 
\begin{equation}
    p(\mathrm{M})=\frac{1}{|\mathcal{K}|}\sum_{\mathrm{k}\in\mathcal{K}}r_\mathrm{k}(\mathrm{M})
\end{equation}
\section{Experiments}
\begin{figure}[t]
    \centering
    \includegraphics[width=0.47\textwidth]{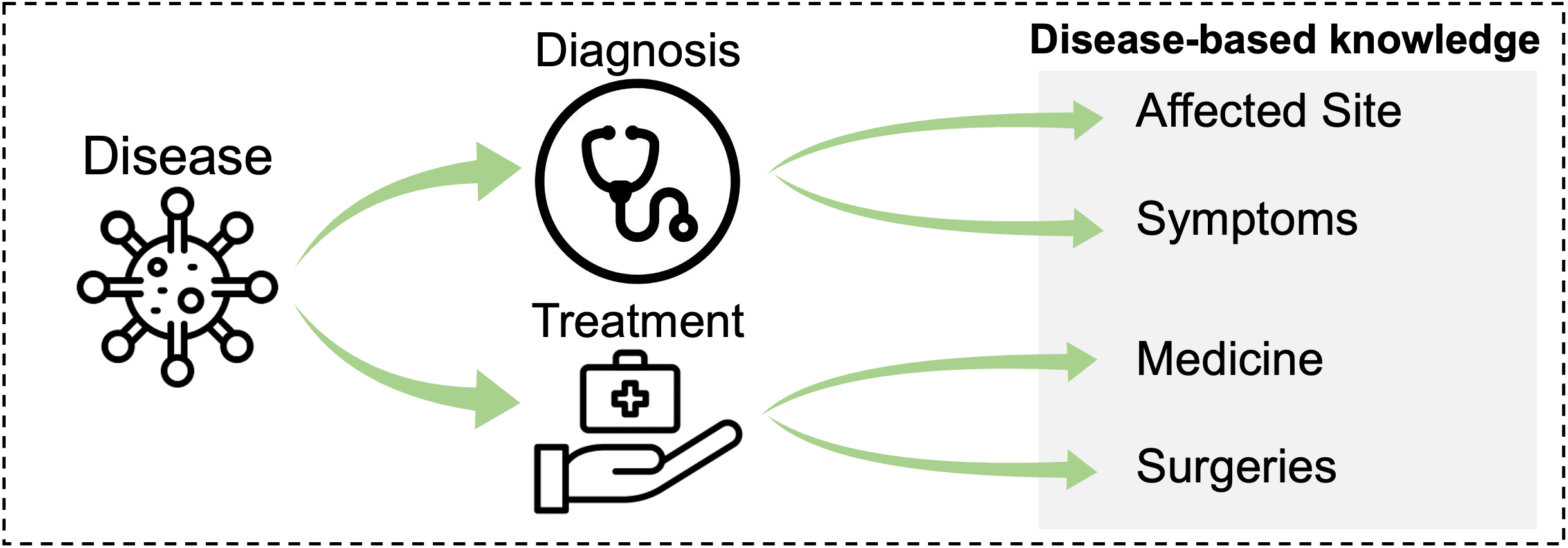}
    \caption{Aspects of disease-related knowledge in DiseK.}
    \label{fig:DiseK}
\end{figure}
\subsection{Experiment Setup}\label{sec:exp setup}
\begin{figure*}[t]
    \centering
    \includegraphics[width=\textwidth]{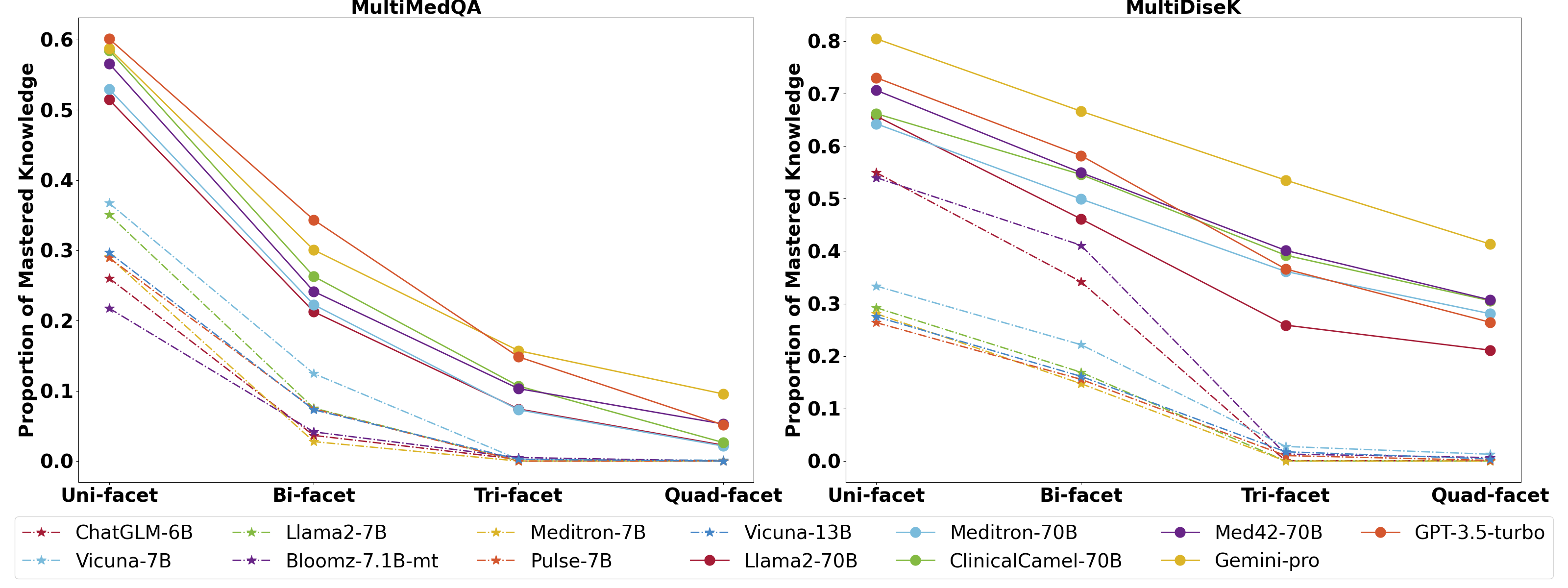}
\caption{Proportion of mastered knowledge points ($p(\mathrm{M})$) evaluated by single-faceted and multi-faceted methods on two datasets. Dash dotted lines refer to LLMs with sizes under 70B, while solid lines denote LLMs larger than 70B. Evaluated facets are added following the sequence: comparison, verification, rectification, and discrimination.}
\label{fig: prop}
\end{figure*}
\paragraph{MultiDiseK Dataset Generation}
Medical knowledge bases explicitly contain knowledge points that can be directly utilized in the proposed method. In this paper, we introduce a disease-centric knowledge base (\textbf{DiseK}) and construct a multifaceted evaluation dataset (\textbf{MultiDiseK}) from it. DiseK is annotated by 20 medical experts for about 3 months. 
It consists of 1,000 common diseases, accompanied by 4 fundamental aspects of medical knowledge (illustrated in Figure \ref{fig:DiseK}). These aspects are closely associated to the clinical decision-making process, involving diagnosis and treatment. Therefore, LLMs must acquire these aspects of medical knowledge to be applicable in clinical decision support systems (CDSS) \cite{wu2018master,liang2019evaluation}. 

The MultiDiseK dataset is constructed based on medical knowledge points in DiseK by using carefully crafted question templates. For the comparison facet, an MCQ (\textbf{4-option}) is generated for each aspect of disease knowledge, where options are formed by selecting attributes that either belong or do not belong to the specified disease. Revision questions are generated by rephrasing each MCQ into two questions, providing either the correct choice or a randomly selected incorrect choice. Multiple-answer questions are generated similarly to MCQs, but with selecting 1-3 attributes as correct options. For true-false questions, we randomly choose an attribute with 50\% probability associated with the disease and 50\% not associated. Participants are then asked to determine whether the given attribute is associated with the disease. For all questions crafted above, we also generate a corresponding negated version by incorporating negation words into the question and modifying the answers correspondingly. This is done to further assess the depth of knowledge mastery by LLMs. Finally, the constructed dataset encompasses a total of 3,167 disease-related knowledge points (some diseases may not have corresponding medications or surgeries), including 6,334 MCQs, 12,668 RQs, 6,334 MAQs, and 6,334 TFQs. More details of this dataset (e.g., question templates, dataset statistics) are presented in Appendix A and B. 
\paragraph{MultiMedQA Dataset Generation}
To make our proposed multifaceted evaluation approach comparable with existing benchmarks, we further construct another dataset \textbf{MultiMedQA} based on a medical benchmark \textbf{MedQA} \cite{jin2021disease}, since several LLMs have achieved notable performance on this benchmark. MedQA is a medical exam dataset that contains \textbf{5-option} multiple-choice questions from the professional medical board exams of different sources. The question in MedQA typically consists of a patient's medical consultation record followed by a question related to the patient's situation (e.g., diagnosis, the next step in management, findings of diagnostic tests). Employing the multifaceted evaluation schema, we rephrase each MedQA question into various types, conducting a multifaceted evaluation of the medical knowledge points embedded in MedQA. To do so, we first selected 800 questions suitable for the multifaceted adaptation from the US exam part (1,273 questions) by regular expressions. After that, we rephrase them into multifaceted questions with heuristic rules. For revision-type questions, we generate them using a method similar to that applied in MultiDiseK construction. For multiple-answer questions, given the challenge of explicitly identifying the knowledge points in the original question, we adopt a solution by retrieving synonyms for the correct option and randomly replacing 0-3 incorrect options with these synonyms to generate a new question. Each question is further paired with a negated version by introducing a negation word in the question. True-false questions were generated by substituting the interrogative word/phrase (e.g., ``which of the following") with either the correct or incorrect option selected from the remaining four options, and negated versions were also created using the same method. The resulting MultiMedQA dataset includes 800 MCQs, 1,600 RQs, 1,600 MAQs, and 3,200 TFQs. More details of MultiMedQA are provided in Appendix C.
\paragraph{Evaluation Setting}
We evaluate LLMs by five-shot learning on the proposed datasets. We report the performance of LLMs under two settings: (1) answer-only \cite{brown2020language}: prompting LLMs with only question-answer pairs; (2) Chain-of-Thought with Self-consistency (CoT+SC) \cite{wang2022self}: prompting LLMs multiple times with question-answer pairs and the chain-of-thoughts, aggregating the results by majority vote to obtain the final answer. For the latter setting, we generate CoTs following the method proposed in \cite{nori2023can} and ask LLMs each question 5 times in our implementation. We only apply the answer-only setting for experiments in MultiDiseK since questions in MultiDiseK do not require sophisticated reasoning in medical cases. We use carefully designed regular expressions to extract answers and observe that they can retrieve answers successfully in most cases. More details are provided in Appendix D.
\begin{table*}[t]
    \begin{subtable}[t]{0.5\textwidth}
    \centering
    \setlength{\tabcolsep}{0.8mm}{

    \begin{tabular}{lccccc}
\hline
\textbf{Model}    & \textbf{Comp.} & \textbf{Rect.} & \textbf{Disc.} & \textbf{Veri.} & \textbf{Average} \\ \hline
Random            & 20.0           & 20.0           & 3.2            & 50.0           & 23.3             \\ \hline
ChatGLM-6B        & 27.7           & 20.3           & 5.7            & 50.6           & 26.1             \\
Vicuna-7B         & 21.0           & 17.7           & 2.1            & 49.4           & 22.6             \\
Llama2-7B         & 20.8           & 23.0           & 0.1            & 49.6           & 23.4             \\
Bloomz-7.1B-mt    & 25.4           & 11.9           & 5.5            & 50.1           & 23.2             \\
Meditron-7B       & 20.6           & 18.8           & 0.0            & 48.9           & 22.1             \\
Pulse-7B          & 19.9           & 14.9           & 0.7            & 49.2           & 21.2             \\
Vicuna-13B        & 20.1           & 17.4           & 0.6            & 51.7           & 22.4             \\ \hline
Llama2-70B        & 41.8           & 30.7           & 10.8           & 54.7           & 34.5             \\
Meditron-70B      & 47.2           & 28.3           & 5.1            & 50.8           & 32.8             \\
ClinicalCamel-70B & 23.9           & 24.9           & 6.4            & 50.8           & 26.5             \\
Med42-70B         & \textbf{59.0}  & 44.8           & \textbf{26.2}  & 57.5           & \textbf{46.9}    \\ \hline
Gemini-pro        & 41.0           & 37.2           & 12.5           & \textbf{59.2}  & 37.5             \\
GPT-3.5-turbo     & 45.5           & \textbf{48.6}  & 12.5           & 58.1           & 41.2             \\ \hline
\end{tabular}}
    \caption{Results in the setting of Answer-only.}
    \label{tab:medqa ao}
  \end{subtable}
  \hfill
  \begin{subtable}[t]{0.5\textwidth}
    \centering
    \setlength{\tabcolsep}{0.8mm}{
        \begin{tabular}{lccccc}
        \hline
\textbf{Model}    & \textbf{Comp.} & \textbf{Rect.} & \textbf{Disc.} & \textbf{Veri.} & \textbf{Average} \\ \hline
Random            & 20.0           & 20.0           & 3.2            & 50.0           & 23.3             \\ \hline
ChatGLM-6B        & 26.0           & 17.2           & 6.8            & 49.1           & 24.8             \\
Vicuna-7B         & 36.7           & 10.8           & 6.2            & 53.0           & 26.7             \\
Llama2-7B         & 35.1           & 18.1           & 5.6            & 51.2           & 27.5             \\
Bloomz-7.1B-mt    & 21.8           & 12.5           & 5.5            & 50.9           & 22.7             \\
Meditron-7B       & 29.1           & 13.3           & 4.8            & 50.2           & 24.3             \\
Pulse-7B          & 28.9           & 19.6           & 5.3            & 50.5           & 26.1             \\
Vicuna-13B        & 29.7           & 15.8           & 5.7            & 52.1           & 25.8             \\ \hline
Llama2-70B        & 50.6           & 35.4           & 10.1           & 58.1           & 38.5             \\
Meditron-70B      & 53.1           & 33.9           & 10.8           & 56.4           & 38.5             \\
ClinicalCamel-70B & 58.2           & 37.5           & 12.0           & 61.3           & 42.2             \\
Med42-70B         & 56.9           & 34.3           & 22.6           & 59.1           & 43.2             \\ \hline
Gemini-pro        & 59.4           & 40.2  & \textbf{34.5}  & \textbf{64.2}  & \textbf{49.6}    \\
GPT-3.5-turbo     & \textbf{60.4}  & \textbf{50.1}  & 22.4           & 62.2           & 48.8             \\ \hline   
        \end{tabular}}
        \caption{Results in the setting of Chain-of-Thought Self Consistency.}
        \label{tab:medqa cotsc}
  \end{subtable}
  \caption{Five-shot accuracies on the \textbf{MultiMedQA} dataset across comparison (Comp.), rectification (Rect.), discrimination (Disc.), and verification (Veri.) capabilities. ``Average" column denotes the macro average of accuracies across all facets.}
    \label{tab:medqa res}
\end{table*}
\begin{table}[t]
\centering
\setlength{\tabcolsep}{0.8mm}{
\begin{tabular}{lccccc}
\hline
\textbf{Model}    & \textbf{Comp.} & \textbf{Rect.} & \textbf{Disc.} & \textbf{Veri.} & \textbf{Average} \\ \hline
Random                          & 25.0                            & 25.0                            & 6.7                             & 50.0                            & 26.7                              \\ \hline
ChatGLM-6B                      & 35.2                            & 27.7                            & 18.5                            & 52.8                            & 33.5                              \\
Vicuna-7B                       & 29.5                            & 24.9                            & 14.0                            & 55.1                            & 30.9                              \\
Llama2-7B                       & 27.8                            & 25.5                            & 15.7                            & 55.2                            & 31.0                              \\
Bloomz-7.1B-mt                  & 34.0                            & 21.0                            & 17.6                            & 53.3                            & 31.5                              \\
Meditron-7B                     & 27.4                            & 25.2                            & 12.5                            & 50.1                            & 28.8                              \\
Pulse-7B                        & 26.1                            & 22.2                            & 2.8                             & 52.9                            & 26.0                              \\
Vicuna-13B                      & 25.6                            & 25.6                            & 9.2                             & 52.5                            & 28.2                              \\ \hline
Llama2-70B                      & 65.6                            & 47.5                            & 33.7                            & 58.1                            & 51.2                              \\
Meditron-70B                    & 66.3                            & 50.3                            & 38.7                            & 63.1                            & 54.6                              \\
ClinicalCamel-70B               & 66.8                            & 63.1                            & 37.0                            & 68.8                            & 58.9                              \\
Med42-70B                       & 72.5                            & 57.3                            & 37.1                            & 64.4                            & 57.8                              \\ \hline
Gemini-pro                      & \textbf{81.7}  & \textbf{72.6}  & \textbf{55.0}  & \textbf{77.0}  & \textbf{71.6}    \\
GPT-3.5-turbo                   & 74.1                            & 59.1                            & 44.8                            & 63.7                            & 60.4                              \\ \hline
\end{tabular}}
\caption{Five-shot accuracies on the \textbf{MultiDiseK} dataset.}
    \label{tab:dkb res}
\end{table}
\paragraph{Metrics}
We employ \textbf{proportion of mastered knowledge points} ($p(\mathrm{M})$ in Sec.\ref{sec:framework}) to measure the overall performance. For each facet, we employ accuracy ($\frac{\text{\#correctly answered questions}}{\text{\#all of the questions}}$) as the fine-grained metric. For MAQs, correct predictions require an exact match with the ground truth answers. Regarding RQs, correctness is determined when both the veracity of the chosen option and the original question's answer are accurately predicted. We observe that some LLMs ``fortunately" achieve high accuracies on RQs by always staying consistent with the provided option since half of the RQs provide the correct options. Therefore, we revise the calculation of revision questions' performance to reduce this bias: $\operatorname{acc}=\frac{1}{N_o}\operatorname{acc}_{T}+\frac{N_o-1}{N_o}\operatorname{acc}_{F}$, where $N_o$ is the number of options, $\operatorname{acc}_{T}$ and $\operatorname{acc}_{F}$ are the accuracies of RQs that provide correct and incorrect options, respectively. The accuracy calculated above is proven to reduce the impact of this bias, and we provide the corresponding proof in Appendix E.
\paragraph{Baseline Models}
We evaluate a total of 13 LLMs with varying sizes in this paper: (1) general LLMs: ChatGLM (6B) \cite{du2022glm}, Llama2 (7B,70B) \cite{touvron2023llama}, Vicuna (7B,13B) \cite{zheng2023judging}, Bloomz-mt (7.1B) \cite{muennighoff-etal-2023-crosslingual}, GPT-3.5-turbo \cite{ouyang2022training} and Gemini-pro \cite{team2023gemini}; (2) medical LLMs: Pulse (7B) \cite{pulse2023}, Meditron (7B,70B) \cite{chen2023meditron}, ClinicalCamel (70B) \cite{toma2023clinical}, and Med42 (70B) \cite{med42}. We have not evaluated GPT-4 \cite{openai2023gpt4} and MedPaLM \cite{singhal2023large}, since GPT-4 is too expensive and MedPaLM is not publicly available yet.
\subsection{Results}
\begin{figure*}[t]
\centering
\begin{subfigure}[t]{0.49\textwidth}
    \centering
    \includegraphics[width=\textwidth]{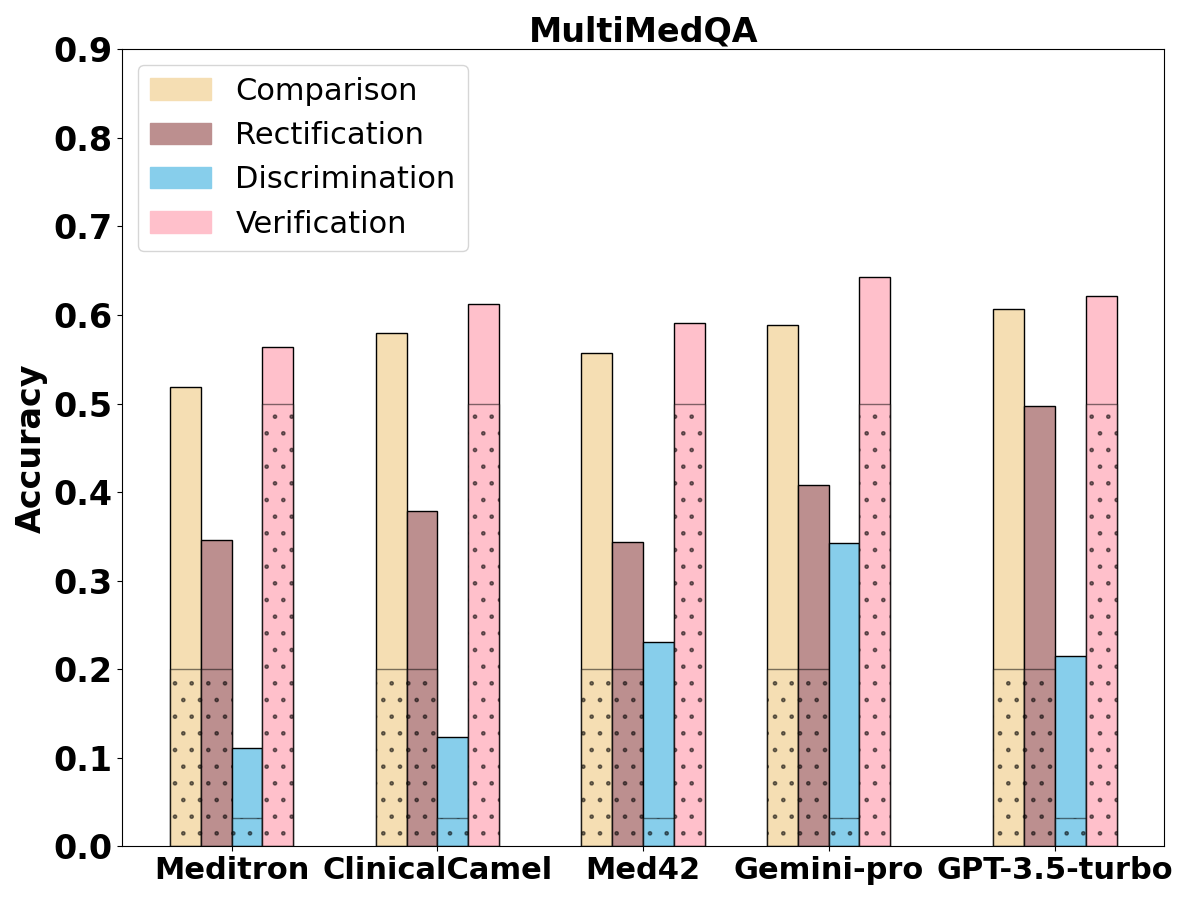}
    \caption{Results on the MultiMedQA dataset.}
    \label{fig:medqa multiface}
\end{subfigure}
\hfill
\begin{subfigure}[t]{0.49\textwidth}
    \centering
    \includegraphics[width=\textwidth]{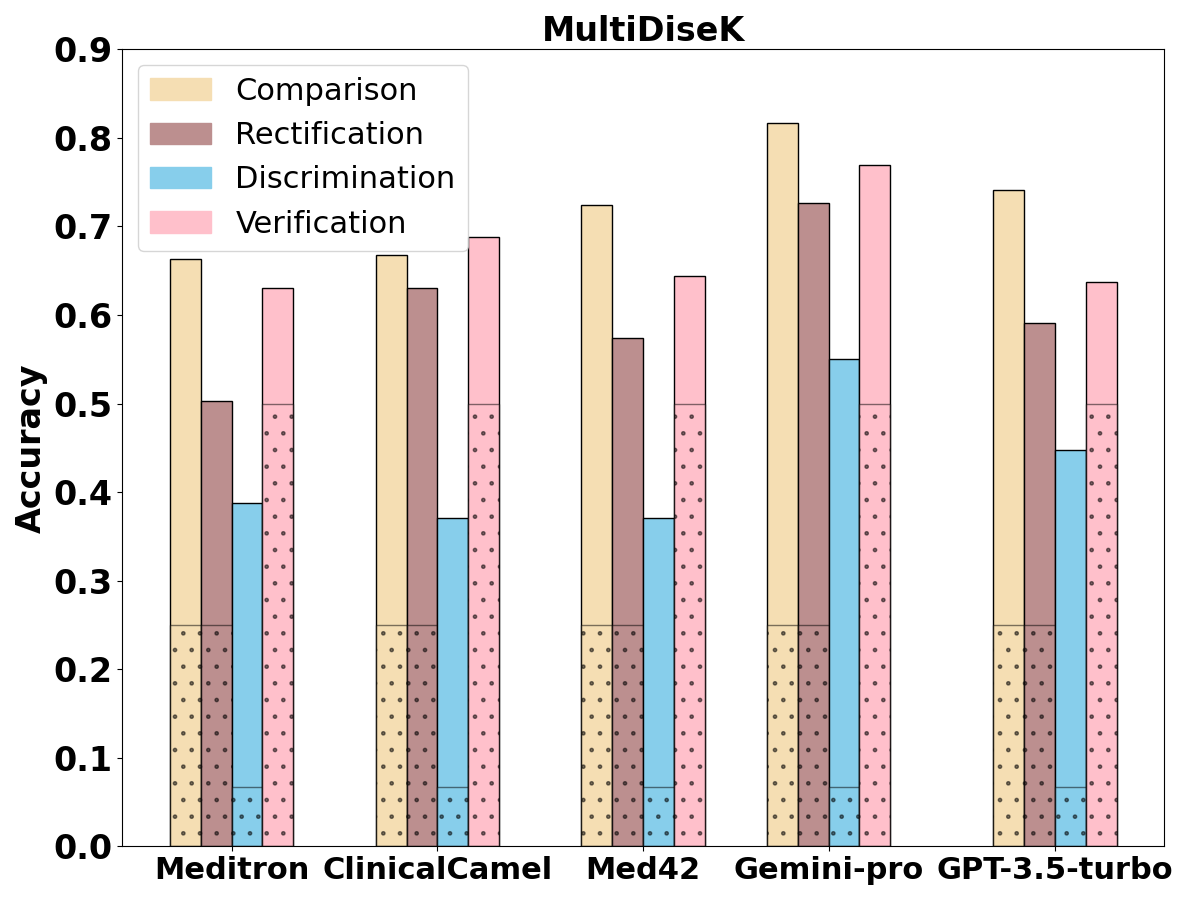}
    \caption{Results on the MultiDiseK dataset.}
    \label{fig:dkb multiface}
\end{subfigure}
\caption{Multifaceted performance of LLMs across the evaluated facets on MultiMedQA and MultiDiseK. Hatched bars: random guessing performance. Solid bars above the hatched part: LLMs gain over random guessing. Meditron, ClinicalCamel, and Med42 are all 70B versions.}
\label{fig: multiface}
\end{figure*}
\paragraph{Single-faceted vs. Multi-faceted}
We first compare the proposed multifaceted evaluation with the conventional single-faceted evaluation. Figure \ref{fig: prop} illustrates the proportion of mastered knowledge points ($p(\mathrm{M})$) by LLMs on the proposed MultiMedQA and MultiDiseK datasets, evaluated using both single-faceted (comparison-type) and multifaceted methods\footnote{Since current LLMs struggle to correctly answer both an affirmative question and its negation simultaneously, we remove negated questions in this analysis to ensure the visibility.}. These LLMs are reported to achieve high performance on existing medical benchmarks, including MedQA. 
We report the performance on the MultiMedQA achieved by the CoT+SC setting since these LLMs generally achieve higher performance in this setting. The experimental results indicate that all LLMs above 70B have effectively mastered a considerable number of knowledge points when evaluated solely from the comparison facet (i.e., \textbf{the original MedQA questions}), consistent with their reported performance on existing benchmarks. However, we observe a \textbf{sharp decline} in the proportion of mastered knowledge points across various LLMs as the number of evaluated facets increases. For example, GPT-3.5-turbo's performance evaluated by 4 facets is around 50\% lower on MultiMedQA and 40\% lower on MultiDiseK compared with the single-faceted results.
Moreover, we observe that though several smaller LLMs (dash dotted lines) also perform notably under single-faceted evaluation, their performance nearly approaches zero when evaluated by $\geq 3$ facets. In contrast, larger LLMs master more knowledge under multifaceted evaluation. We also study different sequences of adding evaluation facets in Appendix F and observe that the conclusions remain consistent. 
The results imply that current LLMs lack a comprehensive mastery of medical knowledge. 
\paragraph{Comparison Across LLMs}
Table \ref{tab:medqa res} and \ref{tab:dkb res} compare LLMs performance across various datasets and settings. LLMs generally perform better on the MultiDiseK dataset since the questions do not involve analysis of specific medical cases. \textbf{Gemini-pro} achieves the highest performance on both datasets with 49.6 and 71.6 in average accuracy, respectively. GPT-3.5-turbo performs similarly to Gemini-pro on MultiMedQA (48.8) but significantly lags behind Gemini-pro on MultiDiseK (60.4). The discrepancy may be attributed to the broader coverage of disease knowledge by Gemini-pro compared with GPT-3.5-turbo, while its ability to apply medical knowledge in specific medical cases is similar to GPT-3.5-turbo. For open-source LLMs in 70B size, we find that several medical LLMs (Med42, ClinicalCamel) significantly surpass their base model LLama2-70B and achieve comparable performance compared to GPT-3.5-turbo on both datasets (46.9 for Med42 on MultiMedQA and 58.9 for ClinicalCamel on MultiDiseK). LLMs that are not larger than 13B perform only slightly better than random guessing. However, they achieve significantly higher performance in the comparison facet and perform similarly or even worse than random guessing on facets such as verification and rectification. One possible explanation is that these two facets represent higher-level capabilities that can manifest only in LLMs with larger sizes. Notably, the comparison-type questions in MultiMedQA are directly sourced from the MedQA dataset. In our study, the performance of GPT-3.5-turbo on this facet (\textbf{60.4}) aligns closely with the reported performance in \cite{nori2023can} (\textbf{60.2}), which could indicate the reliability of our findings. Comparing the Answer-only setting with the CoT+SC setting, we find that larger models significantly benefit more from CoT+SC (except Med42). The effect of CoT+SC varies across facets: for Gemini-pro, CoT+SC largely improves its performance in the comparison (+18.4) and discrimination (+22.0) facets, while it has a limited effect on verification (+5.0) and rectification facets (+3.0). 
\paragraph{Comparison Across Multiple Facets}
We further compare the top-5 LLMs' performance across different capabilities facets in Figure \ref{fig: multiface}. The performance on the MultiMedQA is reported under the CoT+SC setting as well. Note that the hatched bars represent the random guessing performance of the corresponding question type. The experimental results demonstrate that the evaluated LLMs typically exhibit the most significant improvement over random guessing in the \textbf{comparison facet}, followed by the rectification and discrimination facets, and lastly, the verification facet. The high performance on the comparison facet may be caused by the fact that current LLMs have seen more comparison-type questions (MCQs) in their training data to perform well on existing benchmarks. Rectification-type questions are more challenging than comparison-type questions because they require LLMs to determine the correctness of the provided answer and to revise it accurately. Discrimination-type questions also perform worse than comparison-type questions, probably because of their demand for LLMs to discern nuances between concepts instead of merely selecting the most suitable choice. Verification-type questions exhibit the lowest gain, likely due to the need for direct verification based on medical knowledge without additional information from options.

\section{Conclusion and Discussion}
In this paper, we propose a multifaceted evaluation approach, MultifacetEval, designed to probe the actual mastery of medical knowledge by current LLMs. Following this methodology, we construct two multifaceted evaluation datasets, MultiDiseK and MultiMedQA. 
The experimental results demonstrate that current LLMs' medical knowledge mastery is significantly lower than their performance on medical benchmarks suggests, indicating that the proposed MultifacetEval framework offers a more comprehensive assessment of LLMs' medical knowledge mastery.
Furthermore, LLMs demonstrate significant variations in performance across different evaluation facets. These results suggest that \textbf{Current LLMs generally lack a deep, precise, and comprehensive mastery of medical knowledge}, which is the probable cause of the disparity between high performance on medical benchmarks and insufficient performance on real medical scenarios.  
Moreover, although some smaller LLMs are reported to achieve performance comparable to larger LLMs on several benchmarks, they achieve much lower performance on multifaceted datasets, indicating that their mastery of medical knowledge is not as comprehensive as that of larger LLMs.

The above conclusion also provides insights into the development of medical foundation models: (1) Medical foundation models need to be sufficiently large to master medical knowledge comprehensively, deeply, and precisely; (2) Their training should cover a diverse range of medical tasks rather than being restricted to specific ones, making them truly applicable in real-world scenarios.

Finally, it is worth noting that our study is only a first step in exploring the actual mastery of medical knowledge by LLMs. In the future, we plan to evaluate LLMs across additional facets relevant to real medical applications and expand the scale of knowledge points for evaluation, continuously enhancing the comprehensiveness, professionalism, and robustness of the proposed method.
\section*{Acknowledgments}
The work is supported by National Key R\&D Program of China (2021ZD0113402). We thank the anonymous reviewers for helpful comments and feedback.
\bibliographystyle{named}
\bibliography{ijcai24}
\end{document}